\documentclass[twocolumn,letterpaper]{article}
\usepackage{preprint}
\usepackage[utf8]{inputenc}
\usepackage[T1]{fontenc}
\usepackage{amsmath,amssymb,graphicx,booktabs,float,microtype}
\usepackage[numbers,square]{natbib}
\usepackage{titlesec}
\titlespacing{\section}{0pt}{12pt plus 3pt minus 3pt}{1pt plus 1pt minus 1pt}
\titlespacing{\subsection}{0pt}{10pt plus 3pt minus 3pt}{1pt plus 1pt minus 1pt}
\titlespacing{\subsubsection}{0pt}{8pt plus 3pt minus 3pt}{1pt plus 1pt minus 1pt}
\usepackage{titling}
\usepackage{placeins,balance,ragged2e,xurl}
\usepackage[colorlinks=true,linkcolor=purple,urlcolor=blue,citecolor=cyan,
            anchorcolor=black,breaklinks=true]{hyperref}

\newcommand{\method}{EviSlide-MIL}
\newcommand{\R}{\mathbb{R}}
\newcommand{\Author}[2]{\textbf{#1}\textsuperscript{#2}}

\title{EviSlide-MIL: Multi-Query Pooling and Shared Prediction\\
       for Class-Incremental Whole-Slide Image Classification}
\author{%
  \Author{Zhouhan Qian}{1}\quad \Author{Sicheng Chen}{2}\quad
  \Author{Chad Wong}{2}\quad \Author{Tianyi Zhang}{3}\\[0.25em]
  \Author{Enhui Chai}{4}\quad \Author{Zeyu Liu}{4}\quad \Author{Fei Xia}{2}%
}
\date{
  \textsuperscript{1}Duke University \qquad
  \textsuperscript{2}University of California, Irvine\\
  \textsuperscript{3}National University of Singapore \qquad
  \textsuperscript{4}PuzzleLogic\\
  [1em]
  \footnotesize Zhouhan Qian and Sicheng Chen contributed equally to this work. \\
  \footnotesize \textbf{Corresponding author:} Fei Xia\texttt{<fei.xia@uci.edu>} \\
}
\hypersetup{
  pdftitle={EviSlide-MIL: Multi-Query Pooling and Shared Prediction for Class-Incremental Whole-Slide Image Classification},
  pdfauthor={Zhouhan Qian, Sicheng Chen, Chad Wong, Tianyi Zhang, Enhui Chai, Zeyu Liu, Fei Xia}
}

\begin{document}
\twocolumn[\begin{@twocolumnfalse}
  \maketitle
  \thispagestyle{empty}
  \begin{abstract}
Histopathological analysis of whole-slide images (WSIs) plays a critical role in cancer diagnosis, and multiple instance learning (MIL) is widely adopted for WSI analysis under weak supervision. As new tumor subtypes are introduced, extending existing MIL models requires learning new classes while retaining knowledge of previously learned ones. This presents a class-incremental learning challenge, where models must acquire new classes with limited access to historical training data and recognize all learned classes without task identifiers at inference. A major obstacle is forgetting, as updates to feature aggregation and prediction modules can degrade performance on previously learned classes. To address this challenge, we propose \textbf{EviSlide-MIL}, a class-incremental WSI framework that integrates multiple learnable queries per class with a shared prediction module. The queries aggregate frozen patch features into class-specific representations, while the shared prediction module limits parameter growth across incremental stages. By freezing queries associated with previously learned classes, EviSlide-MIL preserves their attention patterns for identical inputs, while experience replay and knowledge distillation further mitigate forgetting as shared parameters are updated. Experiments on eight tumor subtypes from The Cancer Genome Atlas demonstrate that EviSlide-MIL achieves improved performance and reduced forgetting compared with competing methods across most learning settings, demonstrating its potential for scalable and knowledge-preserving WSI classification. Code and pretrained models will be released.
\end{abstract}

  \vspace{0.35cm}
\end{@twocolumnfalse}]

\section{Introduction}

\begin{figure}[t]
    \centering
    \includegraphics[width=\linewidth]{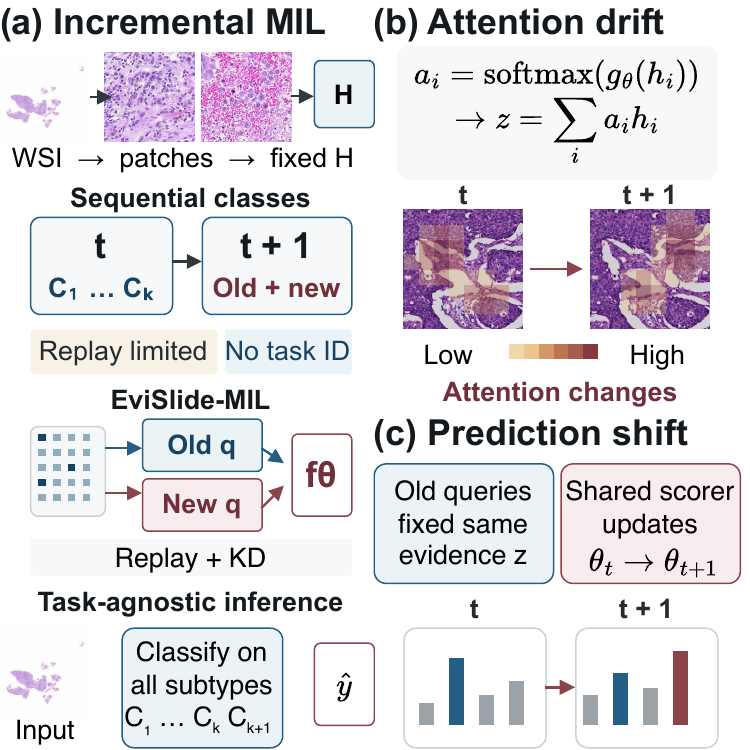}
    \caption{\textbf{Motivation.} (a) Class-incremental WSI classification with limited replay. (b) Updating attention can alter aggregated evidence. (c) Updating the shared predictor can change class scores.}
    \label{fig:motivation}
\end{figure}

Whole-slide image (WSI) analysis is crucial in computational pathology, enabling automated tumor classification from digitized histopathology slides. Due to the gigapixel resolution of WSIs and the limited availability of detailed annotations, multiple instance learning (MIL) has become a widely adopted approach for WSI analysis~\cite{ilse2018attention,lu2021clam}. MIL models typically aggregate features extracted from tissue patches to identify discriminative patterns under slide-level supervision. However, as new tumor subtypes are introduced, conventional MIL models often require retraining on previously collected data. In practice, storing and revisiting all historical patient data may be impractical due to storage and privacy constraints. This motivates class-incremental learning, where models sequentially acquire new classes with limited access to earlier data while recognizing all previously learned classes without task identifiers at inference.

Class-incremental learning must acquire new knowledge while retaining previously learned classes~\cite{rebuffi2017icarl}. Existing continual MIL methods mitigate catastrophic forgetting through experience replay and knowledge distillation~\cite{huang2023conslide,li2025advancing}. However, WSI classification requires preserving both tissue evidence aggregation and class discrimination as new tumor subtypes emerge (Fig.~\ref{fig:motivation}). Although class-specific attention and query-based aggregation offer flexible representations~\cite{lu2021clam,gou2025queryable}, three challenges remain.

\noindent \textit{\textbf{Instability of class-specific evidence aggregation.}}
MIL models rely on attention-based aggregation to identify informative tissue regions for slide-level predictions. During incremental learning, updating shared aggregation parameters for new tumor subtypes can alter the attention assigned to patches associated with earlier classes. Previously learned representations may drift even when the underlying patch features remain unchanged. Although attention distillation can constrain such changes~\cite{li2025advancing}, maintaining class-specific evidence while learning discriminative representations for new subtypes remains an important challenge.

\noindent \textit{\textbf{Interference in shared class prediction.}}
Preserving earlier attention patterns does not necessarily preserve classification performance. Even with fixed patch aggregation, updating the prediction function to accommodate new classes can change the scores assigned to previously learned subtypes. Moreover, introducing new candidate classes increases competition during inference, potentially reducing the relative confidence of earlier classes. This distinction between aggregation and prediction makes forgetting difficult to address through attention preservation alone.

\noindent \textit{\textbf{Balancing incremental adaptability and parameter efficiency.}}
A class-incremental MIL framework should accommodate an expanding set of tumor subtypes without excessive model growth. Assigning independent aggregation and prediction modules to each class can increase parameter requirements as new classes arrive. Conversely, extensively sharing parameters across classes may introduce interference between previously learned and newly acquired knowledge. Limited access to historical patient data further complicates this trade-off, motivating architectures that support class-specific representation learning while maintaining a compact shared prediction mechanism.

To address these challenges, we propose \textbf{EviSlide-MIL}, a class-incremental WSI framework designed to learn new tumor subtypes while preserving previously acquired knowledge and controlling parameter growth. The framework integrates class-specific evidence aggregation, parameter-efficient prediction, and knowledge retention to support incremental learning.

Our main contributions are summarized as follows:

\begin{itemize}
    \item We propose a decoupled architecture that employs \textbf{multi-query class-specific aggregation} to extract evidence from frozen patch features, with a \textbf{shared prediction module} to score the resulting class representations.

    \item To mitigate representation drift and parameter growth, we freeze queries associated with previously learned classes. Each new class introduces only its own queries and a class-specific bias, preserving historical attention patterns for identical inputs while limiting additional parameters.

    \item To balance adaptation and retention, we integrate \textbf{experience replay and knowledge distillation}. Replayed patch features maintain supervision for previously learned classes, while distillation encourages consistency with earlier predictions and reduces forgetting.

    \item Experiments on eight tumor subtypes from The Cancer Genome Atlas~\cite{tcga2013pancancer} demonstrate improved classification performance and reduced catastrophic forgetting across most evaluated settings. Additional experiments with GigaPath features~\cite{xu2024gigapath} examine query multiplicity, prediction sharing, query freezing, and retention strategies.
\end{itemize}

\section{Related Work}
\subsection{Multiple Instance Learning}
Multiple instance learning (MIL) predicts bag-level labels from sets of instances without instance-level annotations~\cite{dietterich1997solving}. ABMIL replaces fixed pooling with learned instance weights~\cite{ilse2018attention}. CLAM introduces class-specific attention and instance clustering~\cite{lu2021clam}, while TransMIL models morphological and spatial dependencies among patches through Transformer-based aggregation~\cite{shao2021transmil}. PANTHER models distributions of morphological prototypes for unsupervised slide representation learning~\cite{song2024panther}. More recently, pathology foundation models such as CONCH and GigaPath provide transferable representations for downstream analysis~\cite{lu2024conch,xu2024gigapath}. These advances improve feature extraction and bag aggregation, but do not directly address incremental learning of new tumor subtypes.

\subsection{Query Pooling}
Query pooling aggregates variable-sized sets of input features into compact representations using learnable queries. Deep Sets establishes permutation-invariant set representations through shared transformations and sum pooling~\cite{zaheer2017deepsets}. Set Transformer introduces pooling by multihead attention with learnable seed vectors~\cite{lee2019set}, while Perceiver uses latent queries to aggregate high-dimensional inputs~\cite{jaegle2021perceiver}. Query2Label employs label queries to extract class-related features through cross-attention, followed by class-specific prediction heads~\cite{liu2021query2label}. For incremental WSI classification, QPMIL-VL combines queryable prototype aggregation with enhanced class-text representations~\cite{gou2025queryable}.

\subsection{Continual Learning}
Continual learning aims to acquire new classes while mitigating catastrophic forgetting, often through experience replay and knowledge distillation~\cite{li2018lwf,rebuffi2017icarl}. Bias correction addresses predictions skewed toward newly introduced classes~\cite{wu2019bic}. For WSI analysis, ConSlide combines hierarchical modeling with breakup--reorganize rehearsal~\cite{huang2023conslide}. AKD preserves attention knowledge, while PMP retains informative patches for replay~\cite{li2025advancing}. More recent approaches include KMP-MIL, which employs parameterized knowledge memory~\cite{yuan2026kmp}, and CooMIL, which adopts context-guided prompt learning~\cite{corso2026coomil}. CoMEL extends continual MIL to tumor localization~\cite{lee2025comel}, while MergeSlide merges models with task-to-class prompt-aligned inference~\cite{bui2026mergeslide}.

\begin{figure*}[t]
\centering
\includegraphics[width=\textwidth]{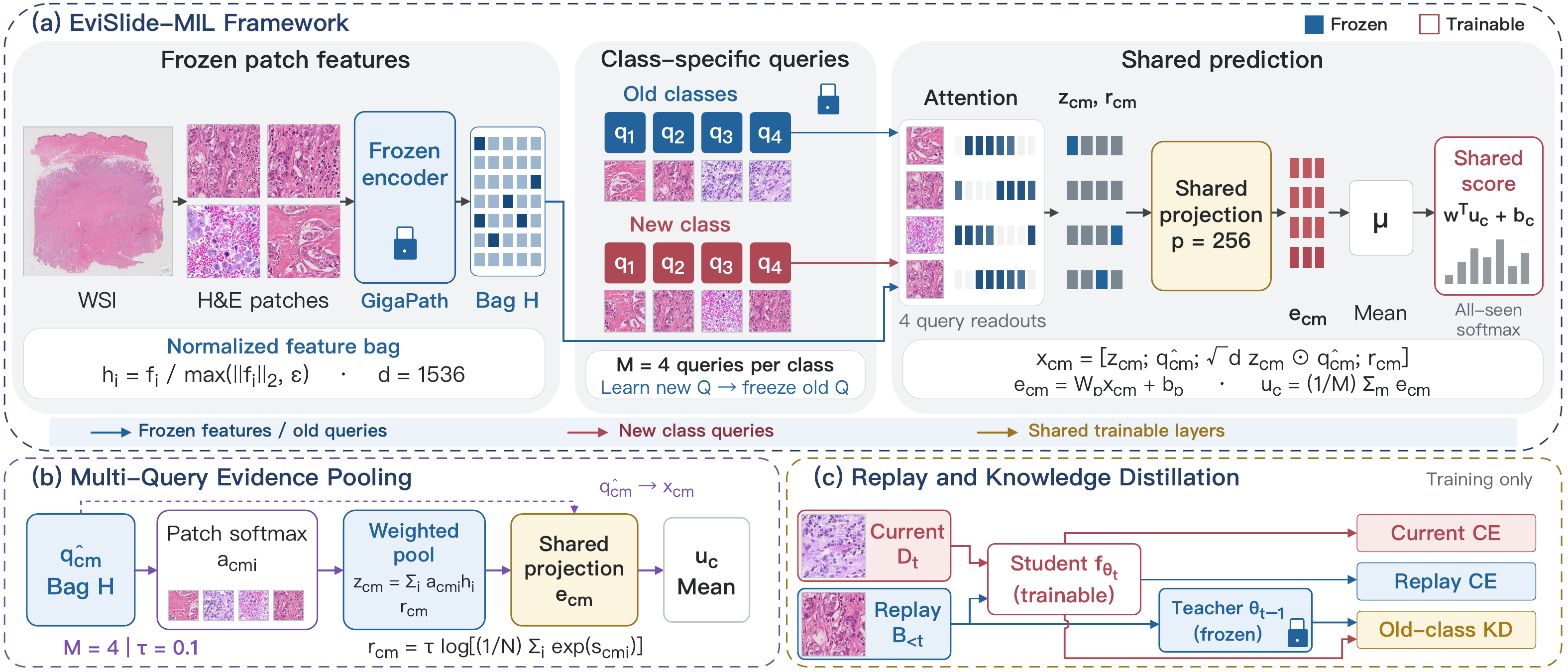}
\caption{\textbf{Overview of EviSlide-MIL.} (a) Four class-specific queries aggregate frozen WSI patch features. Shared projection enables classification across all seen classes. (b) Multi-query pooling extracts patch-level evidence through attention-weighted aggregation. (c) Experience replay and knowledge distillation from a frozen teacher support retention during incremental learning.}
\label{fig:architecture}
\end{figure*}
\section{Method}
\label{sec:method}
\subsection{Overall Structure}
At stage $t$, WSI bags $\mathcal D_t$ introduce classes $\Delta\mathcal C_t$, expanding prediction to $\mathcal C_t=\bigcup_{s\leq t}\Delta\mathcal C_s$. Historical data are restricted to bounded replay memory. Only slide labels supervise aggregation~\cite{li2021dsmil,zhang2022dtfd}. Patch annotations are unnecessary.

\method{} separates class-specific aggregation from shared prediction (Fig.~\ref{fig:architecture}). A frozen encoder supplies patch features, and multiple queries per class produce weighted slide summaries. A shared projection combines each summary with its query and match strength. Mean aggregation, a shared scoring vector and class biases then yield a softmax over all seen classes without organ or task identifiers.

Queries learn at introduction and then freeze. Shared layers and all biases continue adapting through patient replay and a frozen preceding-stage teacher. Replay supervises earlier labels against new competitors. Distillation constrains their relative probabilities.

\subsection{Class-specific Multi-query Pooling}
\label{sec:query-pooling}
A slide supplies frozen features $f_i\in\R^d$, normalized as $h_i=f_i/\max(\|f_i\|_2,\epsilon)$ to form $H=\{h_i\}_{i=1}^{N}$. Each class $c$ has $M$ queries $q_{cm}\in\R^d$, normalized to $\widehat q_{cm}=q_{cm}/\max(\|q_{cm}\|_2,\epsilon)$. At temperature $\tau$, patch attention and pooled features are
\begin{align}
 s_{cmi}&=\widehat q_{cm}^{\top}h_i/\tau,\nonumber\\
 a_{cmi}&=\frac{\exp(s_{cmi})}{\sum_j\exp(s_{cmj})},\qquad
 z_{cm}=\sum_i a_{cmi}h_i.\label{eq:patch-attention}
\end{align}
Each query normalizes over patches independently, yielding its own weighted view without competing for attention with other queries. Pooling is invariant to patch permutation, following the set-based formulation~\cite{zaheer2017deepsets,lee2019set}.

A scalar query--bag compatibility score retains absolute match strength that normalized attention weights discard:
\begin{equation}
 r_{cm}=\tau\left[\log\sum_i\exp(s_{cmi})-\log N\right].
 \label{eq:patch-evidence}
\end{equation}
Subtracting $\log N$ cancels uniform repetition of every patch, but does not remove sensitivity to the selected tissue regions. Both $z_{cm}$ and $r_{cm}$ precede any trainable shared transformation, enabling the stability property below.

A shared projection combines pooled content, query identity, their scaled interaction and match strength:
\begin{align}
 x_{cm}&=[z_{cm};\widehat q_{cm};\sqrt d\,z_{cm}\odot\widehat q_{cm};r_{cm}],\nonumber\\
 e_{cm}&=W_p x_{cm}+b_p,\qquad
 u_c=\frac{1}{M}\sum_{m=1}^{M}e_{cm}.\label{eq:query-embedding}
\end{align}
Here $\odot$ denotes elementwise multiplication and $W_p\in\R^{p\times(3d+1)}$. Mean aggregation weights the projected query representations equally. Queries may overlap spatially, as no diversity constraint assigns a distinct morphological role to each of the learned queries.

\subsection{Shared Prediction and Parameter Growth}
\label{sec:shared-prediction}
A shared vector $w\in\R^p$ and class bias $b_c$ produce
\begin{equation}
 \ell_c=w^{\top}u_c+b_c,\qquad
 p_t(c\mid H)=\frac{\exp(\ell_c)}{\sum_{k\in\mathcal C_t}\exp(\ell_k)}.
 \label{eq:class-score}
\end{equation}
Queries and biases provide class specificity, and the remaining parameters are shared. Adding a class introduces only $Md+1$ parameters, giving the implemented predictor size
\begin{equation}
 P(K)=K(Md+1)+(3d+3)p
 \label{eq:parameter-count}
\end{equation}
for $K$ seen classes, including frozen queries but excluding the encoder. Attention costs $O(NKMd)$, so compact parameter growth does not imply constant inference cost.

Both shared layers are linear. For $\bar x_c=M^{-1}\sum_m x_{cm}$, their composition is
\begin{equation}
 \ell_c=(W_p^{\top}w)^{\top}\bar x_c+w^{\top}b_p+b_c.
 \label{eq:affine-score}
\end{equation}
The common offset cancels in softmax. Projection width therefore changes the parameterization and optimization of an affine scorer, while query-dependent pooling remains nonlinear. Parameter counts and controls retain the separate implemented layers, rather than a collapsed scoring vector.

\subsection{Stable Class Expansion}
\label{sec:class-expansion}
Let $\mathcal P_t^c$ be current patients of class $c$, and $\mathcal S(u)$ the slides of patient $u$. The operator $\operatorname{Low}^{\pi}_M$ returns up to $M$ entries in ascending seeded-priority order. Patient and slide priorities initialize query sources as
\begin{align}
 U_c&=\operatorname{Low}^{\pi_p}_M(\mathcal P_t^c),\qquad L_c=|U_c|,\nonumber\\
 u_{cm}&=U_c[1+((m-1)\bmod L_c)],\nonumber\\
 s_{cm}&=\arg\min_{s\in\mathcal S(u_{cm})}\pi_s(s).
 \label{eq:query-sources}
\end{align}
A private CPU stream samples a nonzero patch from each selected bag:
\begin{align}
 I^+(s)&=\{i:\|f_i^{(s)}\|_2>0\},\nonumber\\
 i_{cm}&\sim\operatorname{Unif}(I^+(s_{cm})),\qquad
 q_{cm}^{\mathrm{init}}=h_{i_{cm}}^{(s_{cm})}.
 \label{eq:query-initialization}
\end{align}
Thus, distinct patients supply queries when available, with cyclic reuse otherwise. All-zero bags are rejected. Controls share candidate bags and patch draws.

Writing $Q_c=(q_{c1},\ldots,q_{cM})$ and $b(c)$ for class introduction, expansion preserves old queries and updates only
\begin{align}
 Q_c^{(t)}&=Q_c^{(b(c))},\qquad c\in\mathcal C_{t-1},\nonumber\\
 \Theta_t^{\mathrm{train}}&=\{W_p,b_p,w\}\cup\{b_c:c\in\mathcal C_t\}\nonumber\\
 &\quad\cup\{Q_c:c\in\Delta\mathcal C_t\}.
 \label{eq:trainable-expansion}
\end{align}
For old classes with identical selected features, normalization and temperature, Eqs.~\ref{eq:patch-attention}--\ref{eq:patch-evidence} then give
\begin{equation}
 (a_c,z_c,r_c)^{(t)}(H)=(a_c,z_c,r_c)^{(b(c))}(H).
 \label{eq:readout-invariance}
\end{equation}
This identity also fixes $\bar x_c$ for old classes, and changing the encoder or selected patches invalidates it. However, with $v_t=(W_p^{(t)})^{\top}w^{(t)}$, the log odds between old classes $c,k\in\mathcal C_{t-1}$ can still change:
\begin{align}
 &\log\frac{p_t(c\mid H)}{p_t(k\mid H)}
 -\log\frac{p_{t-1}(c\mid H)}{p_{t-1}(k\mid H)}\nonumber\\
 &\quad=(v_t-v_{t-1})^{\top}(\bar x_c-\bar x_k)\nonumber\\
 &\qquad+(b_c^{(t)}-b_c^{(t-1)})-(b_k^{(t)}-b_k^{(t-1)}).
 \label{eq:prediction-drift}
\end{align}
New classes also change the global argmax, requiring prediction-level retention beyond fixed attention.

\subsection{Replay-supervised Incremental Learning}
\label{sec:replay-training}
For $K=|\mathcal C_t|$ seen classes and patient budget $B$, class $c$ receives
\begin{equation}
 n_c=\lfloor B/K\rfloor+\mathbf1\{j(c)\leq B\bmod K\},
 \label{eq:memory-quota}
\end{equation}
where $j(c)$ is its introduction-order index. Let $V_{t-1}^c$ denote retained historical patient identifiers. Replay priority $\rho_p$ selects stored and current patients for the labeled memory:
\begin{align}
 V_t^c&=\operatorname{Low}^{\rho_p}_{n_c}(V_{t-1}^c\cup\mathcal P_t^c),\nonumber\\
 I_u&\sim\operatorname{Unif}\{I\subseteq[N_{s(u)}]:|I|=R\},\nonumber\\
 \widetilde H_u&=\{h_i^{(s(u))}:i\in I_u\},\nonumber\\
 \mathcal B_t&=\{(\widetilde H_u,c):u\in V_t^c,\ c\in\mathcal C_t\}.
 \label{eq:memory-admission}
\end{align}
Here $[N]=\{1,\ldots,N\}$, $s(u)$ is the patient's lowest-priority slide, and $R$ is the stored patch count. Patch sampling applies only to new entries, historical entries retain their features. Infeasible quotas, insufficient patches and violations of the serialized byte cap are rejected without rereading discarded slides.

Let $\mathcal D_t^c$ and $\mathcal B_t^c$ contain bags with label $c$. Independent streams sample with replacement, defining current and replay laws $\mu_t$ and $\nu_t$:
\begin{align}
 c&\sim\operatorname{Unif}(\Delta\mathcal C_t),\quad
 H\sim\operatorname{Unif}(\mathcal D_t^c),\nonumber\\
 \widetilde c&\sim\operatorname{Unif}(\mathcal C_{t-1}),\quad
 \widetilde H\sim\operatorname{Unif}(\mathcal B_t^{\widetilde c}).
 \label{eq:balanced-sampling}
\end{align}
Only earlier classes supply replay losses. The teacher $\theta_{t-1}$ is frozen before expansion and receives the same replay bag as student $\theta_t$. Their old-class probabilities are renormalized at temperature $T_d$~\cite{hinton2015distilling}:
\begin{equation}
 p_{\theta}^{\mathrm{old}}(c\mid H;T_d)=
 \frac{\exp(\ell_{\theta,c}(H)/T_d)}
 {\sum_{k\in\mathcal C_{t-1}}\exp(\ell_{\theta,k}(H)/T_d)}.
 \label{eq:old-class-distillation}
\end{equation}
With $\mathcal L_{\mathrm{CE}}(H,y)=-\log p_t(y\mid H)$ over all seen classes, the three unit-weight objectives are
\begin{align}
 \mathcal L_t={}&\mathbb E_{\mu_t}\mathcal L_{\mathrm{CE}}(H,c)
 +\mathbb E_{\nu_t}\mathcal L_{\mathrm{CE}}(\widetilde H,\widetilde c)\nonumber\\
 &+T_d^2\mathbb E_{\nu_t}
 D_{\mathrm{KL}}\!\left(p_{\theta_{t-1}}^{\mathrm{old}}(\widetilde H;T_d)
 \,\|\,p_{\theta_t}^{\mathrm{old}}(\widetilde H;T_d)\right).
 \label{eq:training-loss}
\end{align}
Stage one omits both replay expectations. Replay CE compares old labels against new competitors, while KD preserves relative old-class probabilities~\cite{rebuffi2017icarl,buzzega2020der}. Inference uses neither memory nor the teacher. Training budgets and numerical settings appear in Sec.~\ref{sec:experiments}.

\begin{table*}[!t]
\centering
\caption{\textbf{Development system comparisons.} Means over 16 conditions. F1/recall in \%, forgetting in pp. Bold marks each group maximum (minimum for forgetting). Additional inputs: $\dagger$ class text, $\ddagger$ parent context, $\S$ self-supervision.}
\label{tab:main-baselines}
\footnotesize
\setlength{\tabcolsep}{3pt}
\renewcommand{\arraystretch}{1.12}
\begin{tabular*}{\textwidth}{@{\extracolsep{\fill}}l*{6}{c}@{}}\toprule
& \multicolumn{6}{c}{\textit{GigaPath}}\\
\cmidrule(l){2-7}
Metric & \method{} & \shortstack{CLAM-SB\\+ AKD} & \shortstack{CLAM-MB\\+ Replay/KD} & PMP & \shortstack{TransMIL\\+ Replay/KD} & \shortstack{ABMIL\\+ Replay/KD}\\\midrule
Final F1 $\uparrow$ & \textbf{77.80} & 72.68 & 72.62 & 71.50 & 48.00 & 64.02\\
Forgetting $\downarrow$ & \textbf{13.12} & 19.36 & 20.62 & 21.09 & 41.24 & 28.32\\
Old recall $\uparrow$ & \textbf{79.50} & 74.53 & 73.25 & 72.48 & 52.28 & 66.37\\
New recall $\uparrow$ & 93.64 & 94.71 & 95.01 & 94.10 & 92.26 & \textbf{95.12}\\\bottomrule
\end{tabular*}

\vspace{5pt}
\begin{tabular*}{\textwidth}{@{\extracolsep{\fill}}l*{7}{c}@{}}\toprule
& \multicolumn{3}{c}{\textit{Projected CONCH}} & \multicolumn{2}{c}{\shortstack{\textit{Hierarchical CONCH}\\\textit{Fine patches}}} & \multicolumn{2}{c}{\shortstack{\textit{ConvNeXt: fine patches}\\\textit{Epoch training}}}\\
\cmidrule(lr){2-4}\cmidrule(lr){5-6}\cmidrule(l){7-8}
Metric & \method{} & QPMIL-VL$^{\dagger}$ & KMP-MIL$^{\dagger}$ & \method{} & CooMIL$^{\dagger\ddagger}$ & \method{} & ConSlide$^{\ddagger\S}$\\\midrule
Final F1 $\uparrow$ & \textbf{81.10} & 77.70 & 70.37 & 78.12 & \textbf{86.78} & \textbf{54.91} & 44.60\\
Forgetting $\downarrow$ & \textbf{9.71} & 11.32 & 15.34 & 12.42 & \textbf{3.09} & \textbf{21.31} & 38.36\\
Old recall $\uparrow$ & \textbf{83.71} & 79.91 & 74.63 & 81.02 & \textbf{89.43} & \textbf{57.71} & 44.00\\
New recall $\uparrow$ & 92.12 & \textbf{92.60} & 85.29 & \textbf{89.05} & 86.56 & 60.86 & \textbf{80.18}\\\bottomrule
\end{tabular*}
\end{table*}

\section{Experiments}
\label{sec:experiments}
\subsection{Datasets and Evaluation Metrics}
\label{sec:datasets}
We evaluate eight TCGA subtypes~\cite{tcga2013pancancer} in four two-class stages: lung (LUAD/LUSC), breast (IDC/ILC), renal (KIRC/KICH) and esophageal (EAC/ESCC). Reverse order inverts these organ pairs. Training uses slide labels and inherited patient-disjoint training/validation partitions. Each configuration covers folds 2--5, two orders and two prescribed seeds: 16 trajectories over 1,223 distinct validation patients. The held-out comparison uses 379 additional patients (394 slides), disjoint from the full 1,523-patient development pool. Its analysis plan and all 128 stage checkpoints were fixed before test access.

Primary performance is final eight-class patient macro-F1. Old/new recall averages the first six/last two introduced classes. Forgetting is the mean old-class decline from peak post-introduction recall to final recall. For the held-out comparison, 5,000 subtype-stratified patient bootstrap draws are shared across methods and conditions to obtain paired percentile 95\% CIs, conditional on the fitted models.

\subsection{Implementation Details}
\label{sec:implementation}
The default uses $M=4$, $p=256$, $\tau=0.1$, $T_d=2$ and normalization constant $\epsilon=10^{-12}$. GigaPath~\cite{xu2024gigapath} supplies 1,536-dimensional features, capped at 2,048 training and 4,096 evaluation patches by deterministic Hilbert ordering and evenly spaced selection. Queries initialize from four current-class patients. AdamW~\cite{loshchilov2019adamw} resets each stage with weight decay $10^{-4}$. Learning rates are $10^{-4}$ initially. Later, new queries retain that rate while shared parameters and all biases use $3\times10^{-5}$. Each stage performs 1,200 updates with terminal checkpoints, accumulating four current and four eligible replay bags per update. Projected CONCH~\cite{lu2024conch} accumulates sixteen of each. Hierarchical CONCH uses fine features from complete regions. ConvNeXt~\cite{liu2022convnext} uses epoch training with independent validation checkpoint selection.

\begin{table}[!t]
\centering
\caption{\textbf{Held-out TCGA evaluation.} Means over 16 conditions on the same 379 patients. F1/recall in \%, forgetting in pp.}
\label{tab:p0-test}
\footnotesize
\setlength{\tabcolsep}{5pt}
\begin{tabular*}{\columnwidth}{@{\extracolsep{\fill}}lcc@{}}\toprule
Metric & \method{} & \shortstack{ABMIL\\Replay/KD}\\\midrule
Macro-F1 & 75.84 & 63.27\\
Old recall & 77.96 & 66.02\\
New recall & 90.83 & 91.22\\
Forgetting & 16.58 & 29.00\\\bottomrule
\end{tabular*}
\end{table}

\begin{figure}[!t]
\centering
\includegraphics[width=\linewidth]{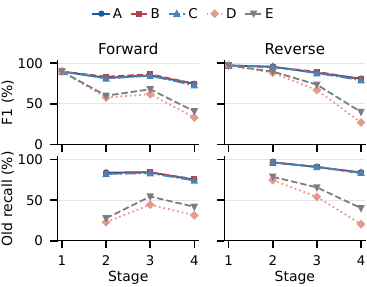}
\caption{\textbf{Development-stage retention.} GigaPath means over eight conditions per order. A--E follow Tab.~\ref{tab:architecture-controls}. Old recall starts at stage two. All controls retain memory and student replay forwards.}
\label{fig:retention-analysis}
\end{figure}

\begin{figure*}[!t]
\centering
\includegraphics[width=\textwidth]{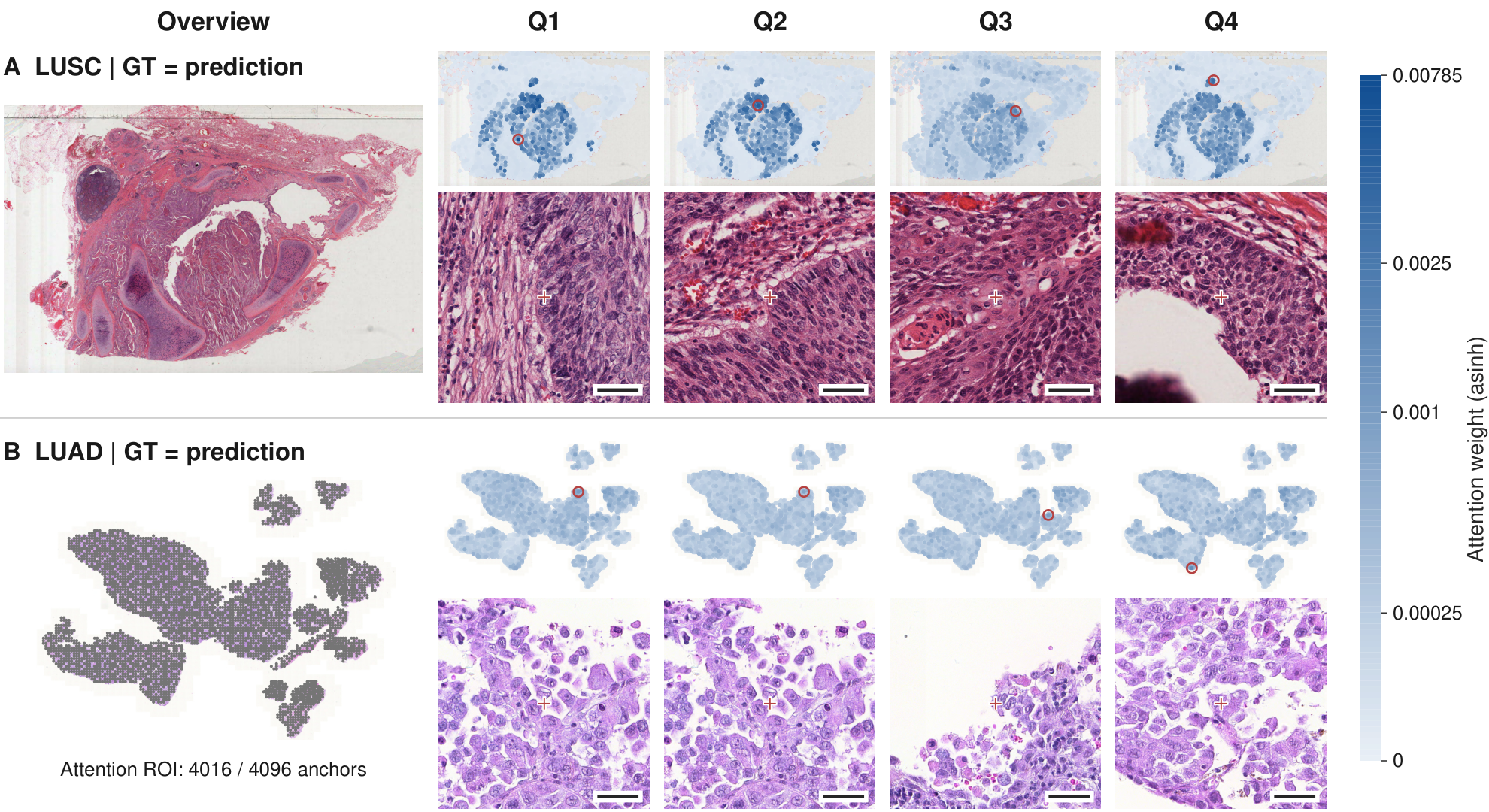}
\caption{\textbf{Four queries in LUSC and LUAD validation cases.} Both cases retain full 4,096-anchor normalization and a shared asinh display scale. Case B displays 4,016 anchors in a tissue-selected ROI without renormalization. Red rings mark unmodified maxima.}
\label{fig:attention-success}
\end{figure*}

\subsection{Comparison with Existing Methods}
Tab.~\ref{tab:main-baselines} compares \method{} with attention-based MIL (ABMIL~\cite{ilse2018attention}, CLAM~\cite{lu2021clam}), TransMIL~\cite{shao2021transmil}, pseudo-bag replay (PMP~\cite{li2025advancing}), vision--language continual MIL (QPMIL-VL~\cite{gou2025queryable}, KMP-MIL~\cite{yuan2026kmp}), context-guided CooMIL~\cite{corso2026coomil} and hierarchical ConSlide~\cite{huang2023conslide}.

\method{} achieved the highest mean final F1 in three of four input/training groups. With GigaPath, it reached 77.80\% F1, outperforming CLAM-SB with AKD~\cite{lu2021clam} by 5.12 pp, while improving old recall by 4.97 pp and reducing forgetting by 6.24 pp. With projected CONCH, it achieved 81.10\% F1, exceeding QPMIL-VL~\cite{gou2025queryable} and KMP-MIL~\cite{yuan2026kmp} by 3.40 and 10.73 pp, respectively, with the strongest old-class retention. Under ConvNeXt training, it surpassed ConSlide~\cite{huang2023conslide} by 10.31 pp in F1 and reduced forgetting by 17.05 pp, despite lower new recall (60.86\% versus 80.18\%). However, CooMIL~\cite{corso2026coomil} led hierarchical CONCH (86.78\% versus 78.12\% F1), leveraging additional parent context, text, task specialists and class masking. In contrast, \method{} scores all seen classes without task identifiers. These results highlight its retention advantage and the trade-off with new-class acquisition.

\subsection{Held-out Classification and Retention}
\label{sec:heldout-results}
The held-out evaluation tests whether the retention advantage extends to unseen patients, a key check on the stability of incremental predictions. On 379 held-out TCGA patients, with the analysis plan and checkpoints fixed before test access, \method{} achieved 75.84\% macro-F1 versus 63.27\% for ABMIL with Replay/KD (Tab.~\ref{tab:p0-test}). The paired gain was 12.57 pp (95\% CI, 10.14--15.16), with positive gains in both orders. Old recall increased from 66.02\% to 77.96\%, while new recall remained similar (91.22\% versus 90.83\%). Forgetting fell by 12.42 pp (95\% CI, 9.90--15.39), linking the improvement mainly to retention of earlier subtypes. Substantial errors remained in vulnerable old classes. Forward-order LUAD/LUSC recalls were 52.38\%/30.00\%, versus 33.93\%/6.67\% for ABMIL; reverse-order ESCC recall was 55.36\% versus 14.29\%. Thus, better aggregate retention did not eliminate subtype-specific forgetting.

 \subsection{Attention Visualization}
We visualize class-specific query attention to examine how multiple queries distribute evidence within a slide, making their aggregation behavior interpretable. Fig.~\ref{fig:attention-success} shows ground-truth-class attention in selected correctly classified LUSC and LUAD cases. In LUSC, Q1, Q2 and Q4 emphasize overlapping central regions, while Q3 spreads attention more broadly. In LUAD, Q1/Q2 share a maximum, whereas Q3/Q4 peak elsewhere. The lower images provide spatial context, revealing variations in cellular density, nuclear morphology and tissue organization across the attended regions. These patterns highlight two complementary behaviors: overlapping queries reinforce shared morphological evidence, while spatially distinct queries capture additional tissue characteristics. Notably, this diversity emerges without explicit constraints on query specialization. By combining shared and complementary attention, multi-query pooling can integrate multiple morphological patterns and better represent tissue heterogeneity within a slide.

\begin{figure}[!t]
\centering
\includegraphics[width=\linewidth]{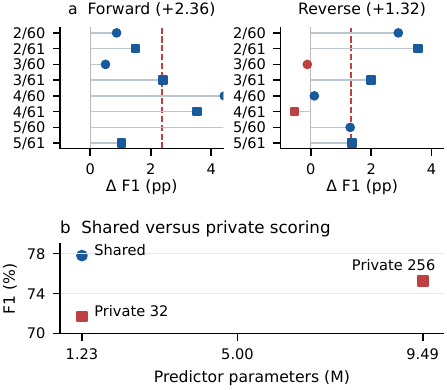}
\caption{\textbf{Query and predictor controls.} (a) Four-query minus single-query F1 in 16 matched conditions. (b) F1 versus predictor size. Counts include frozen queries and exclude the encoder.}
\label{fig:architecture-analysis}
\end{figure}

\section{Discussion}
\subsection{Multi-query Pooling and Shared Prediction}
We first investigate how multi-query pooling and shared prediction contribute to incremental learning performance. Fig.~\ref{fig:architecture-analysis} shows that four queries improved F1 by 1.84 pp over one query, with gains in 14 of 16 conditions, supporting multiple-view aggregation~\cite{gou2025queryable}. Shared prediction reached 77.80\% F1 versus 71.70\% for private width-32 branches at the same final 1.23M-parameter budget. Private width 256 increased size to 9.49M and reached 75.27\%. Together, these results highlight the complementary benefits of multi-query aggregation and shared prediction: the former enriches slide-level representations, while the latter achieves stronger classification with greater parameter efficiency than class-specific predictors.

\subsection{Mean versus Learned Query Aggregation}
\begin{figure}[!t]
\centering
\includegraphics[width=\linewidth]{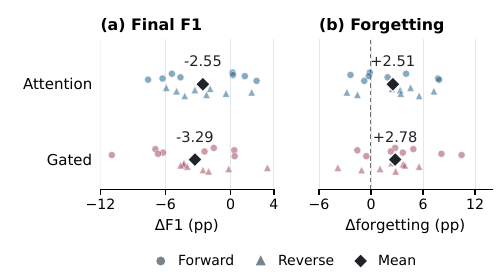}
\caption{\textbf{Learned query aggregation minus mean pooling.} (a) Changes in final F1 score. (b) Changes in forgetting. Both are measured in percentage points (pp). Circles and triangles represent forward and reverse class orders, respectively, while diamonds indicate mean differences. Dashed lines mark zero change.}
\label{fig:query-aggregation-ablation}
\end{figure}

We next investigate whether learned query aggregation improves performance over simple mean pooling. Fig.~\ref{fig:query-aggregation-ablation} shows that attention and gated pooling reduced F1 from 77.80\% to 75.25\% and 74.51\%, respectively, corresponding to decreases of 2.55 and 3.29 pp. Meanwhile, forgetting increased by 2.51 and 2.78 pp, indicating weaker retention of previously learned classes. Together, these results demonstrate the advantage of mean aggregation over learned query weighting, achieving stronger classification and knowledge retention with a simpler aggregation mechanism.

\begin{figure}[t]
\centering
\includegraphics[width=\linewidth]{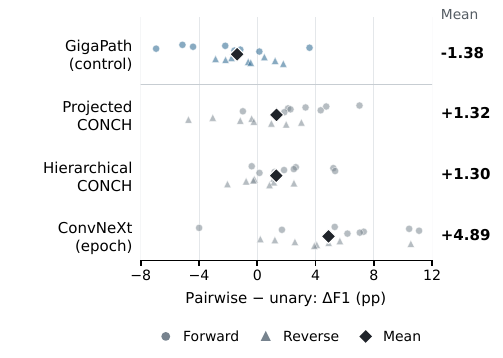}
\caption{\textbf{Pairwise minus unary scoring.} GigaPath is a matched architecture control. Other rows compare separately executed archived versions. Each row contains 16 fold/order/seed pairs. Diamonds show means.}
\label{fig:pairwise-scoring-ablation}
\end{figure}

\subsection{Contribution of Pairwise Scoring}
We further investigate whether pairwise interactions between class representations improve incremental learning performance. Fig.~\ref{fig:pairwise-scoring-ablation} shows that pairwise scoring reduced F1 from 77.80\% to 76.42\% on GigaPath, while increasing forgetting by 1.84 pp and model size from 1.230M to 1.361M parameters. In contrast, it improved F1 by 1.32, 1.30 and 4.89 pp with projected CONCH, hierarchical CONCH and ConvNeXt, respectively. Together, these results suggest that the benefit of pairwise scoring depends on the feature representation and training configuration, while the simpler unary predictor offers better performance and parameter efficiency under the matched GigaPath setting.

\begin{table}[!t]
\caption{\textbf{Freezing and loss controls.} Means over 16 GigaPath conditions. F1/recall are percentages; forgetting is pp. A: default; B: old queries updated; C: no KD; D: no replay CE; E: neither replay loss. Current CE and student replay forwards remain active.}
\label{tab:architecture-controls}\centering\small
\setlength{\tabcolsep}{3pt}
\begin{tabular}{@{}lrrrrr@{}}\toprule
Metric & A & B & C & D & E\\\midrule
Macro-F1 & 77.80 & 77.51 & 75.93 & 30.01 & 40.35\\
Forgetting & 13.12 & 13.43 & 14.40 & 66.85 & 52.13\\
Old recall & 79.50 & 79.34 & 78.79 & 25.84 & 40.91\\
New recall & 93.64 & 93.97 & 91.97 & 94.44 & 94.34\\\bottomrule
\end{tabular}
\end{table}

\subsection{Attention Stability and Predictive Retention}
We next investigate how freezing old queries, knowledge distillation (KD) and replay supervision contribute to knowledge retention. Tab.~\ref{tab:architecture-controls} shows that freezing old queries improved F1 from 77.51\% to 77.80\% and reduced forgetting by 0.31 pp, indicating a modest benefit from stable attention. Removing KD reduced F1 by 1.87 pp and increased forgetting by 1.28 pp. More notably, removing replay CE caused F1 to drop to 30.01\%, with forgetting increasing from 13.12 to 66.85 pp. Removing both replay losses yielded 40.35\% F1 and 52.13 pp forgetting, while new recall remained above 94\% in both settings. Together, these results highlight the importance of replay supervision for preserving earlier classes, with query freezing and KD providing additional stability. Stable attention alone is insufficient to prevent catastrophic forgetting without explicit old-class supervision.

\subsection{Effect of Task Order on Query Freezing}
We further investigate whether class order changes the retention benefit of freezing old queries. Tab.~\ref{tab:query-freezing-orders} shows that freezing improved reverse-order F1 by 0.70 pp and old recall by 1.17 pp, with 1.11 pp less forgetting, whereas continued updating increased new recall by 1.92 pp. In the forward order, updating improved F1 by 0.11 pp and old recall by 0.85 pp but reduced new recall by 1.26 pp. Together, these results show how freezing and continued adaptation shift the balance between old-class retention and new-class acquisition, revealing an order-dependent trade-off that is obscured when results are pooled across class orders.

\begin{table}[!t]
\centering
\caption{\textbf{Query freezing under different task orders.} Means over eight GigaPath conditions per order with replay CE and KD retained. F1/recall are percentages; forgetting and $\Delta$ are pp. $\Delta$ is updating minus freezing, computed before rounding.}
\label{tab:query-freezing-orders}
\small
\setlength{\tabcolsep}{3pt}
\begin{tabular}{@{}llrrr@{}}\toprule
Order & Metric & Frozen & Updated & $\Delta$\\\midrule
Forward & F1 & 74.63 & 74.75 & +0.11\\
 & Forgetting & 16.35 & 15.87 & -0.49\\
 & Old recall & 74.93 & 75.78 & +0.85\\
 & New recall & 98.45 & 97.19 & -1.26\\\midrule
Reverse & F1 & 80.96 & 80.27 & -0.70\\
 & Forgetting & 9.88 & 10.99 & +1.11\\
 & Old recall & 84.08 & 82.91 & -1.17\\
 & New recall & 88.83 & 90.75 & +1.92\\\bottomrule
\end{tabular}
\end{table}

\subsection{Replay Supervision and Distillation}
\begin{figure}[!t]
\centering
\includegraphics[width=\linewidth]{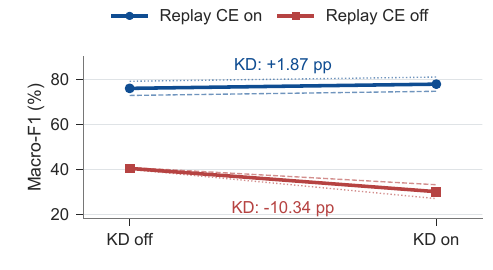}
\caption{\textbf{Replay CE determines the KD effect.} Thick lines average 16 conditions; thin lines show each order.}
\label{fig:replay-kd}
\end{figure}

We finally investigate how replay supervision influences the effectiveness of knowledge distillation (KD) in incremental learning. Fig.~\ref{fig:replay-kd} shows that KD improved F1 by 1.87 pp with replay CE but reduced it by 10.34 pp without replay CE, yielding a 12.21-pp interaction consistent across both class orders. Without replay CE, KD also reduced old recall by 15.07 pp and increased forgetting by 14.72 pp, indicating substantial degradation in knowledge retention. While replay CE directly distinguishes old classes from newly introduced competitors, KD preserves relative probabilities among old classes~\cite{rebuffi2017icarl,buzzega2020der}. Together, these results establish replay CE as the primary component for preserving earlier classes, with KD providing complementary benefits when combined with explicit historical supervision.

\subsection{Limitations and Future Work}
EviSlide-MIL advances class-incremental WSI classification by combining multi-query pooling, parameter-efficient shared prediction, and replay-supervised knowledge retention. Despite these improvements, evaluation remains limited to eight TCGA subtypes and four incremental stages, with performance varying across feature representations and training configurations. The reliance on frozen encoders and stored patient features also limits adaptability and introduces storage and privacy considerations. Future work will explore longer and task-free learning streams~\cite{aljundi2019taskfree}, adaptive feature learning and privacy-aware replay. External multi-institutional validation is needed to assess site-specific shifts~\cite{howard2021histologysignatures} and the balance between knowledge retention and new-class acquisition.

\section{Conclusion}
We presented EviSlide-MIL, a class-incremental WSI classification framework that integrates multi-query pooling, shared prediction, and replay-supervised knowledge retention. EviSlide-MIL achieved the highest final F1 in three of four evaluated settings, demonstrating strong classification performance and reduced catastrophic forgetting. Extensive ablations further demonstrated the benefits of each proposed component. Furthermore, EviSlide-MIL shows the potential for scalable and continuously evolving computational pathology systems, holding broader impact for developing adaptable, efficient pathology AI systems.

\section*{LLM Usage Statement}
We used OpenAI ChatGPT to assist with implementation, experiment orchestration, log and result checks, writing and editing, and figure preparation. The authors take responsibility for the manuscript and its results.

\FloatBarrier
\begingroup
\RaggedRight
\renewcommand{\bibsection}{\section*{\refname}\addcontentsline{toc}{section}{\refname}}
\bibliography{references}
\endgroup
\twocolumn[{\centering
  \pdfbookmark[0]{Supplementary Material}{supplementary-material}%
  \Large\bfseries Supplementary Material\par\vspace{1em}}]
\setcounter{section}{0}
\setcounter{table}{0}
\setcounter{figure}{0}
\setcounter{equation}{0}
\renewcommand{\thesection}{S\arabic{section}}
\renewcommand{\thetable}{S\arabic{table}}
\renewcommand{\thefigure}{S\arabic{figure}}
\renewcommand{\theequation}{S\arabic{equation}}
\renewcommand{\theHsection}{supp.\arabic{section}}
\renewcommand{\theHtable}{supp.\arabic{table}}
\renewcommand{\theHfigure}{supp.\arabic{figure}}
\renewcommand{\theHequation}{supp.\arabic{equation}}

\section{Data and reproducible training protocol}
\label{sec:supp-repro}
\paragraph{Cohort and evaluation units.}
The benchmark uses eight TCGA subtypes: lung adenocarcinoma (LUAD), lung squamous cell carcinoma (LUSC), invasive ductal carcinoma (IDC), invasive lobular carcinoma (ILC), kidney renal clear-cell carcinoma (KIRC), kidney chromophobe carcinoma (KICH), esophageal adenocarcinoma (EAC) and esophageal squamous cell carcinoma (ESCC). Breast and esophageal cases are selected by these diagnoses in the source tables; renal labels follow TCGA-KIRC/KICH membership. Patients are identified by the first three TCGA barcode components, and all slides from a patient form one evaluation unit. Methods share the inherited patient partitions. Before training, we check training/validation disjointness and feature-file availability.

Tab.~\ref{tab:cohort} gives the validation composition within the 1,612-slide development feature pool. Seeds are 60/61 for GigaPath, 62/63 for projected CONCH, 64/65 for hierarchical CONCH and 66/67 for ConvNeXt. Training folds overlap. All five folds informed development. The subsequent frozen evaluation uses the separate held-out cohort described in Sec.~\ref{sec:supp-test}.

\begin{table}[h]
\caption{Validation cohort for the main comparisons. Repeated orders and seeds reuse these patients. Earlier development also used fold 1 for model assessment.}
\label{tab:cohort}\centering\small
\begin{tabular}{lrrrr}\toprule
Subtype & Fold 2 & Fold 3 & Fold 4 & Fold 5\\\midrule
LUAD & 20 & 12 & 19 & 16\\
LUSC & 10 & 14 & 10 & 10\\
IDC & 119 & 127 & 128 & 117\\
ILC & 32 & 34 & 25 & 31\\
KIRC & 83 & 82 & 84 & 81\\
KICH & 18 & 18 & 16 & 17\\
EAC & 10 & 7 & 14 & 11\\
ESCC & 12 & 17 & 14 & 15\\\midrule
Patients & 304 & 311 & 310 & 298\\
Slides & 337 & 332 & 320 & 314\\\bottomrule
\end{tabular}
\end{table}

\paragraph{Feature preparation and bag selection.}
GigaPath bag caps and normalization follow main-text Secs.~\ref{sec:query-pooling} and~\ref{sec:implementation}. Hilbert-based subsampling includes both endpoints; smaller bags retain all patches. Initialization and memory admission use the training cap, and selected patches remain fixed across visits. Checksums identify the cached inputs used in the current comparisons; records of the historical GigaPath image preprocessing are incomplete.

CONCH ViT-B/16 v1.0 extraction uses bicubic resizing, a $448\times448$ center crop, RGB conversion and ImageNet normalization. Projected CONCH uses the contrastive projection and $\ell_2$ normalization, giving 512-dimensional features with the same 2,048/4,096 patch caps. Hierarchical CONCH retains unnormalized 512-dimensional pre-contrastive fine and parent features. Each region supplies an $8\times8$ grid of $512\times512$ fine patches and a parent image resized to $512\times512$. Region extraction targets a $4096\times4096$ field at $0.5\,\mu$m per pixel with half-field stride. A thumbnail mask requires saturation above 0.08 and mean intensity between 20 and 235; accepted regions contain at least 15\% masked tissue. Regions follow a seeded hash order, independent of labels and predictions. For sparse slides, extraction uses a denser stride and, when needed, a halved physical field rescaled to the same pixel dimensions. Flat bags deduplicate overlapping patch origins; hierarchical bags retain complete regions.

Hierarchical training/validation uses at most 32/64 regions. \method{} applies the frozen $512\times512$ contrastive projection from the same CONCH checkpoint, then $\ell_2$ normalizes and flattens the fine features, discarding parent features and grouping. ConvNeXt inputs use torchvision ConvNeXt-Tiny ImageNet-1K V1 weights, ImageNet normalization and the same region/fine-patch geometry, with 768-dimensional outputs; fine patches and parent images are encoded separately. The extraction checkpoints are verified for each feature profile.

\paragraph{Initialization details.}
The seeded patient and slide priorities in main-text Sec.~\ref{sec:class-expansion} use SHA-256. Every control receives a fresh model initialization.

\paragraph{Additional replay and optimization settings.}
The memory rule in main-text Sec.~\ref{sec:replay-training} uses a budget of 60 patients, 64 distinct patches per retained slide, and a 48 MiB serialized-size cap. Independent random generators preserve current and replay sampling streams across controls. Loss sums are divided by the accumulation count.

AdamW uses $(\beta_1,\beta_2)=(0.9,0.999)$ and optimizer $\epsilon=10^{-8}$, distinct from the feature-normalization constant. Learning rates remain constant within each stage. Other default settings are given in main-text Sec.~\ref{sec:implementation}; the ConvNeXt stopping rule appears in Sec.~\ref{sec:supp-comparisons}.

\paragraph{Additional metric conventions.}
Slide probabilities are averaged within each patient before taking the class argmax. Metric definitions follow main-text Sec.~\ref{sec:datasets}. Undefined F1 contributions are zero, and validation stages require support for every seen class. Old recall and forgetting are undefined before the first increment. Introduction recall averages recall on each new pair at its introduction stage; backward transfer averages final minus introduction recall over old classes, in pp. Means weight conditions equally. Development contrasts are descriptive; Sec.~\ref{sec:supp-test} explains the held-out uncertainty estimates.

\section{Baseline inputs and training conditions}
\label{sec:supp-comparisons}
Main-text Tab.~\ref{tab:main-baselines} compares complete systems in four input/training groups, preserving each competitor's architecture and adaptation mechanism. ABMIL and the current GigaPath controls have verified input and sampling records. For historical CLAM, PMP and TransMIL, incomplete feature checksums leave cache identity with the current reference unverified; these comparisons are descriptive. Input information and computation remain method specific even with a shared encoder or current-data budget.

\paragraph{GigaPath architecture and mechanism controls.}
CLAM-SB with AKD, CLAM-MB with replay/KD, TransMIL with replay/KD and ABMIL with replay/KD combine MIL architectures with continual-learning mechanisms. ABMIL uses gated-attention pooling, a width-512 embedding and width-256 attention network, with learning rate $10^{-4}$ throughout. Fixed-update controls use 1,200 updates per stage and terminal checkpoints. TransMIL uses AdamW in place of its published Lookahead--RAdam/epoch recipe. PMP retains pseudo-bag memory selection and proportional independent replay updates; features, patient protocol, optimizer, accumulation, instance supervision and checkpoint selection differ from its published recipe.

\paragraph{Projected CONCH.}
QPMIL-VL and KMP-MIL supplement the shared visual features with frozen text towers, prompts, pools and calibration, without patient replay. KMP-MIL retrieves from its universal class-incremental pool without a true task identifier. Each stage uses 1,200 updates and sixteen current bags per update. \method{} accumulates per-bag gradients with AdamW and patient replay/KD. QPMIL-VL and KMP-MIL use joint minibatch objectives and Adam at 0.001 and 0.005, respectively, with weight decay $5\times10^{-4}$. All use terminal checkpoints.

\paragraph{Hierarchical CONCH.}
CooMIL concatenates raw fine and parent features in its MIL branch and uses normalized projected sums in its text-conditioned branch. It retains frozen specialists, prompts and fourteen training centers per task, predicts a task and then masks classes. \method{} directly scores all seen classes from transformed fine patches and retains patient replay. Both use 1,200 updates with four accumulated current bags per update. CooMIL stores specialists, text and centers as persistent state; its patient-replay wrapper is empty. Context, text, experts, routing and processing change together in these comparisons.

\paragraph{ConvNeXt training and selection.}
ConSlide uses fine and parent features, ten first-task self-supervised epochs, region-buffer augmentation and continuous Adam state. Main-text Tab.~\ref{tab:main-baselines} uses Adam at $10^{-5}$, single-bag updates and 1,100 regions under 256 MiB. \method{} uses fine features, stage-reset AdamW, four-bag accumulation and its 60-patient memory under 48 MiB. The final accumulation batch is normalized by its actual size.

Both methods use complete-epoch current-slide shuffles and at most fifty supervised epochs, with independent stopping and checkpoint selection. Selection minimizes all-seen-class slide cross-entropy on the current-stage validation cohort from zero-based epoch five. Ties save the model and reset patience; a worse result stops training when patience reaches ten or the epoch exceeds ten. ConSlide restores the best model while retaining terminal Adam moments. Independent stopping and selection permit different training durations and exposure despite the shared sequence of early epochs.

\section{Definitions of the main ablations}
\label{sec:supp-nopairwise}
\paragraph{Selection and reference model.}
We removed the pairwise branch after the completed 144-trajectory development study. The 160-trajectory follow-up and the later 48-trajectory loss matrix were each fixed on 3 November 2026, before their respective training. References comprise sixteen verified default runs and sixteen original pairwise mean-model runs. The default was fixed for both follow-up studies.

\paragraph{Query and aggregation controls.}
The single query is initialized as the normalized mean of the same four unit patch candidates used by the default. Class expansion uses the same candidate draws, while the initialization distribution and parameter count change. The old-query update control continues training old queries at $10^{-4}$ while retaining the default's remaining settings, including Replay CE and KD.

\paragraph{Sharing and parameter budget.}
The private controls replace both the shared projection and unary vector with a separate branch for each class. New branches copy a fixed seed-specific initialization template. With eight classes, input dimension $d=1536$ and four queries, private width $p$ gives
\begin{equation}
P_{\mathrm{private}}(p)=8\bigl[4d+1+(3d+3)p\bigr].
\end{equation}
Width 32 was chosen from this budget before validation. The match is in final parameter count, with differences in intermediate-stage capacity, template storage, initialization and computation.

The affine-score derivation in main-text Sec.~\ref{sec:shared-prediction} also applies to private branches: any positive width represents the same effective linear scoring vector. These controls vary parameterization, optimization and budget while retaining separate layers and AdamW updates.

\begin{figure}[t]
\centering
\includegraphics[width=\linewidth,trim=30bp 0bp 0bp 138.5bp,clip]{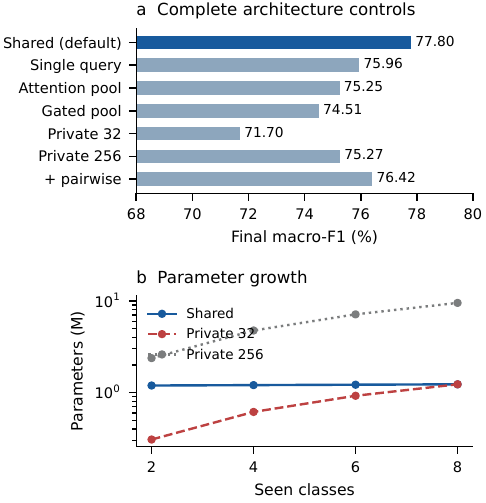}
\caption{\textbf{Predictor parameter growth.} Counts include frozen queries and exclude the encoder. Shared and Private 32 match at the final stage but have different intermediate-stage budgets and initialization.}
\label{fig:supp-architecture-analysis}
\end{figure}

Fig.~\ref{fig:supp-architecture-analysis} shows parameter growth across stages.

\paragraph{Pairwise reference.}
The $+$pairwise row restores the shared comparison network:
\begin{align}
G_{ck}&=g([u_c;u_k;u_c-u_k;u_c\odot u_k]),\nonumber\\
\ell_c&=w^{\top}u_c+b_c+
\frac{1}{K-1}\sum_{k\ne c}(G_{ck}-G_{kc}),
\label{eq:supp-pairwise-reference}
\end{align}
where $u_c$ represents class $c$, $K$ is the number of seen classes and $g$ is a shared $4p\rightarrow128\rightarrow1$ MLP with GELU. The comparison term is zero for $K=1$. All other settings follow the default configuration.

Tab.~\ref{tab:supp-architecture-controls} supplies additional architecture-control metrics.

\begin{table}[!htbp]
\caption{\textbf{Additional architecture-control metrics.} Means over 16 conditions; final F1 is reported in the main-text Discussion. Recall is in \%, forgetting in pp, and size is predictor parameters in millions. All controls retain freezing and all losses.}
\label{tab:supp-architecture-controls}\centering\footnotesize
\setlength{\tabcolsep}{3pt}
\begin{tabular}{@{}lrrrr@{}}\toprule
Configuration & Forget. & Old & New & Size\\\midrule
Single query & 14.34 & 78.20 & 91.34 & 1.193\\
Attn. pooling & 15.63 & 77.05 & 90.83 & 1.246\\
Gated pooling & 15.90 & 76.38 & 88.06 & 1.263\\
Private 256 & 15.43 & 77.42 & 93.64 & 9.492\\
Private 32 & 17.42 & 75.04 & 91.85 & 1.230\\
$+$ pairwise & 14.96 & 77.21 & 89.87 & 1.361\\
\bottomrule
\end{tabular}
\end{table}

\paragraph{Loss-control implementation.}
The replay CE$\times$KD controls retain the default architecture, initialization, optimizer and current data. Replay admission, memory contents and old-bag sampling remain active in every cell. A zero replay CE coefficient retains its student forward and multiplies the supervised loss by zero. With KD off, no teacher is constructed or executed. These controls therefore remove supervision without removing the stored features or student replay computation.

\section{Analyses supporting the main results}
\label{sec:supp-support}
\begin{figure}[!htbp]
\centering
\includegraphics[width=\linewidth,trim=0bp 268bp 0bp 0bp,clip]{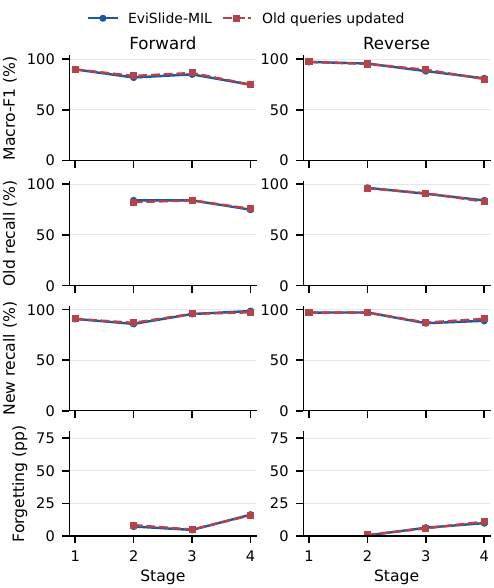}
\par\vspace{1pt}
{\scriptsize\hspace*{.18\linewidth}\makebox[.30\linewidth]{Forward}\hspace{.18\linewidth}\makebox[.30\linewidth]{Reverse}\par}
\includegraphics[width=\linewidth,trim=0bp 0bp 0bp 141bp,clip]{figures/supplement/query_freezing.pdf}
\caption{\textbf{Additional query-freezing trajectories.} New recall and forgetting over four stages, averaging eight conditions per order. The controls share architecture, initialization and all losses. Forgetting is undefined at stage one.}
\label{fig:nopairwise-freezing}
\end{figure}

\paragraph{Additional freezing trajectories.}
Main-text Tab.~\ref{tab:query-freezing-orders} reports final order-specific outcomes. Fig.~\ref{fig:nopairwise-freezing} adds the trajectories of new recall and forgetting; F1 and old-recall trajectories are shown in main-text Fig.~\ref{fig:retention-analysis}.

\begin{figure}[!htbp]
\centering
{\scriptsize
\textcolor[HTML]{185A9D}{\rule{8pt}{.6pt}}\;A\quad
\textcolor[HTML]{B5434C}{\rule{8pt}{.6pt}}\;B\quad
\textcolor[HTML]{4684B9}{\rule{8pt}{.6pt}}\;C\quad
\textcolor[HTML]{E69A8D}{\rule{8pt}{.6pt}}\;D\quad
\textcolor[HTML]{7B7D7E}{\rule{8pt}{.6pt}}\;E\par}
\vspace{2pt}
{\scriptsize\hspace*{.18\linewidth}\makebox[.30\linewidth]{Forward}\hspace{.18\linewidth}\makebox[.30\linewidth]{Reverse}\par}
\includegraphics[width=\linewidth,trim=0bp 0bp 0bp 173bp,clip]{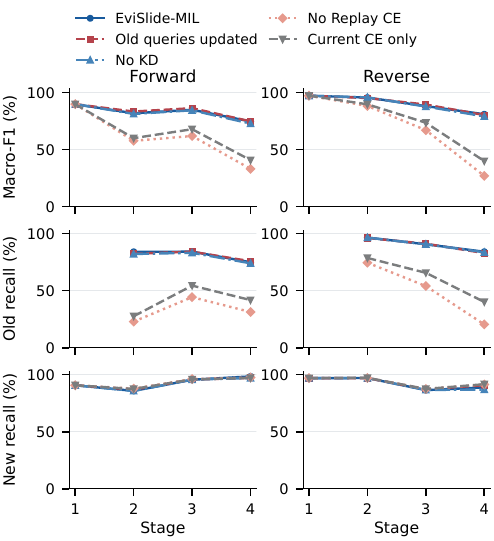}
\caption{\textbf{Stage-wise acquisition of new classes.} New recall covers the current pair, averaged over eight conditions per order. A--E follow main-text Tab.~\ref{tab:architecture-controls}. All controls retain memory and student replay forwards.}
\label{fig:supp-stage-performance}
\end{figure}

\paragraph{Additional acquisition trajectories.}
Fig.~\ref{fig:supp-stage-performance} adds new-class recall trajectories for the five freezing and loss controls in main-text Tab.~\ref{tab:architecture-controls}.

\paragraph{Order-specific loss outcomes and contrasts.}
Tab.~\ref{tab:supp-nopairwise-loss-orders} reports loss-off outcomes separately by order; the full-supervision reference appears in main-text Tab.~\ref{tab:query-freezing-orders}. For replay CE coefficient $R$ and KD coefficient $K$, let $m_{RK}$ denote a metric. The six predeclared contrasts are $m_{11}-m_{10}$, $m_{01}-m_{00}$, $m_{11}-m_{01}$, $m_{10}-m_{00}$, $m_{11}-m_{00}$ and
\begin{equation}
I=(m_{11}-m_{10})-(m_{01}-m_{00}).
\end{equation}
Their pooled values are recoverable from main-text Tab.~\ref{tab:architecture-controls}; all 96 condition-level contrast records remain in the source CSV. Positive F1/recall and negative forgetting favor the first configuration.

\begin{table}[!htbp]
\caption{\textbf{Additional loss-off outcomes by task order.} Eight conditions per order. Columns indicate (replay CE, KD); the $(1,1)$ reference is in main-text Tab.~\ref{tab:query-freezing-orders}. F1/recall are percentages and forgetting is pp.}
\label{tab:supp-nopairwise-loss-orders}\centering\footnotesize
\setlength{\tabcolsep}{3pt}
\begin{tabular}{@{}llrrr@{}}\toprule
Order & Metric & (1,0) & (0,1) & (0,0)\\\midrule
Forward & F1 & 72.78 & 33.09 & 40.77\\
 & Forgetting & 18.00 & 60.25 & 50.27\\
 & Old recall & 73.97 & 31.28 & 41.74\\
 & New recall & 97.04 & 97.51 & 97.06\\\midrule
Reverse & F1 & 79.08 & 26.93 & 39.93\\
 & Forgetting & 10.81 & 73.46 & 53.99\\
 & Old recall & 83.62 & 20.40 & 40.07\\
 & New recall & 86.89 & 91.38 & 91.61\\
\bottomrule\end{tabular}
\end{table}

\begin{figure}[!htbp]
\centering
\includegraphics[width=\linewidth,trim=0bp 230bp 0bp 0bp,clip]{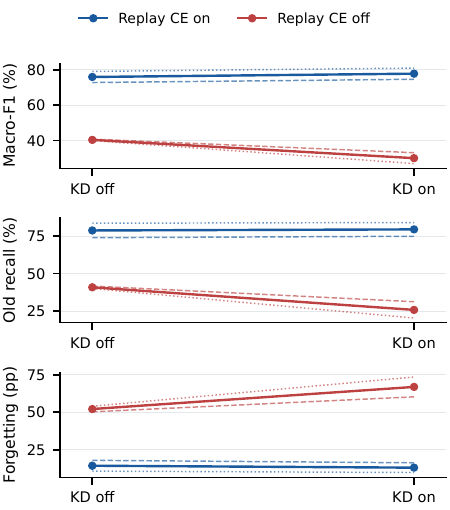}
\par\vspace{1pt}
\includegraphics[width=\linewidth,trim=0bp 0bp 0bp 99bp,clip]{figures/supplement/replay_kd.pdf}
\caption{\textbf{Additional replay/KD readouts.} Old recall and forgetting complement the F1 analysis in main-text Fig.~\ref{fig:replay-kd}. Thick curves average 16 conditions; dashed/dotted curves show forward/reverse means. KD-off omits the teacher; memory and student replay forwards remain active.}
\label{fig:supp-replay-kd}
\end{figure}

\paragraph{Additional retention readouts.}
Fig.~\ref{fig:supp-replay-kd} extends the replay/KD analysis to old recall and forgetting.

\paragraph{Historical-version provenance.}
The version comparison is reported in main-text Fig.~\ref{fig:pairwise-scoring-ablation}. The archived CoMEL-condition reference remains separate from the ordinary GigaPath reference, even where their reported means coincide. Execution records are preserved for both model versions.

\section{Resources and source data}
\label{sec:supp-source-data}
Predictor parameter counts exclude the frozen encoder. State storage additionally includes fixed transformations and initialization templates where present; teacher, optimizer, replay and CUDA allocation are recorded separately. Default replay occupies approximately 22.59 MiB on GigaPath, 7.59 MiB on CONCH and 11.34 MiB on ConvNeXt, under the 48 MiB cap. Competitor text, expert, center and region resources contribute separate persistent costs.

The CSV files in \texttt{anc/source\_data} contain 208 new trajectories/832 stages and 32 reused references/128 stages, including individual fold/order/seed outcomes, class recalls, resources, source-summary checksums, matched differences and all 96 condition-level loss contrasts. The README maps files to tables and defines resource timers. Missing telemetry is blank. Timings measure predictor training on shared hardware and exclude encoder extraction.

\section{Extended protocol and analysis figures}
\label{sec:supp-extended-analysis}
\begin{figure*}[!t]
\centering
\includegraphics[width=\textwidth]{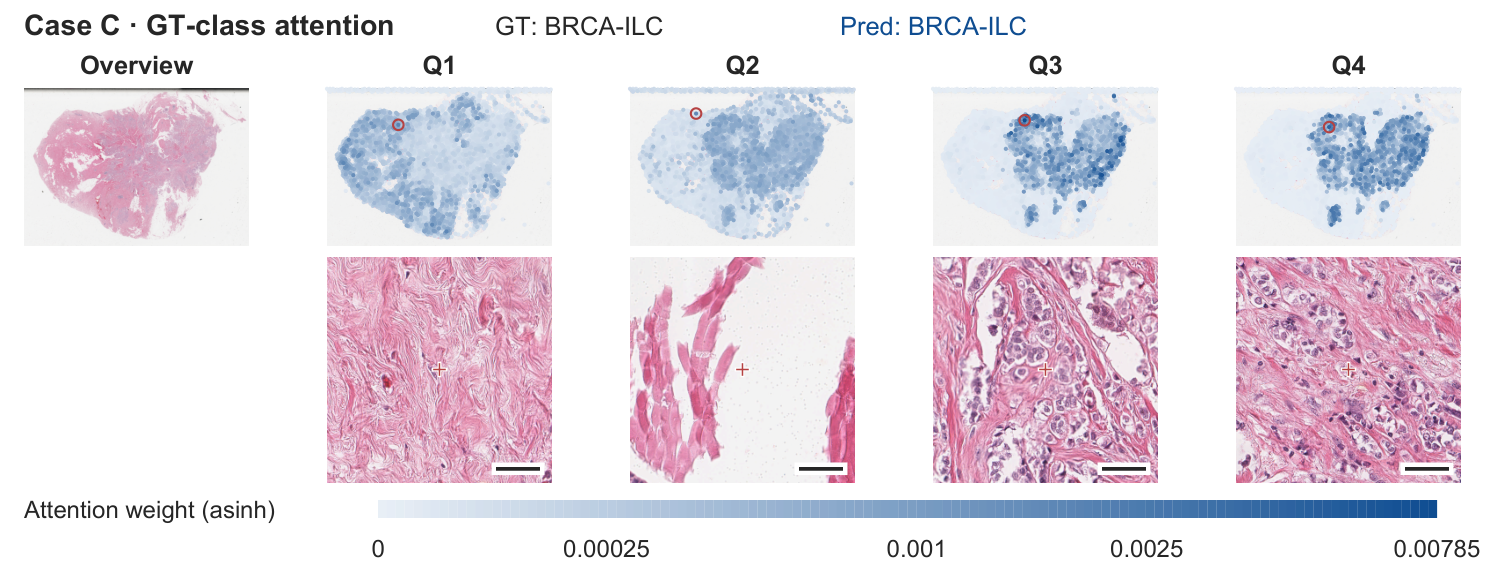}
\caption{\textbf{Additional ground-truth-class attention example: correctly classified ILC.} The main paper shows the LUSC and LUAD cases; this panel retains the same four-query layout and shared asinh display scale with raw-attention ticks. Weights retain full 4,096-anchor normalization. Red rings mark unmodified argmax coordinates, including repeats. Lower images show 256\,$\mu$m spatial contexts with unverified correspondence to historical feature tiles; scale bars are 50\,$\mu$m. These cases visualize model attention; annotation-based tumor-localization accuracy was not evaluated.}
\label{fig:supp-attention-success}
\end{figure*}

\paragraph{Additional attention example.}
Fig.~\ref{fig:supp-attention-success} provides an additional ILC example alongside the main-text LUSC and LUAD cases.

The quantitative figures summarize development-validation conditions, with patients reused across orders and seeds. Sec.~\ref{sec:supp-nopairwise} defines the controls, Sec.~\ref{sec:supp-support} gives supporting contrasts, and Sec.~\ref{sec:supp-test} reports held-out results.

\begin{figure}[!htbp]
\centering
\includegraphics[width=\linewidth]{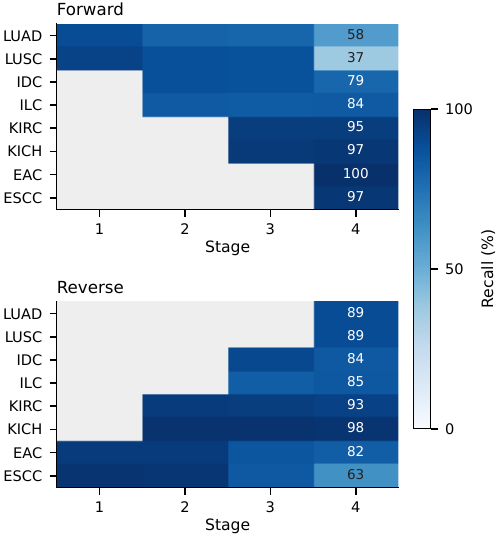}
\caption{\textbf{Development subtype recalls.} Means over eight conditions per order. Both panels retain the same biological subtype order and 0--100\% color scale; gray denotes unseen classes. Final-stage values are annotated, and orders reuse validation patients.}
\label{fig:supp-class-recall}
\end{figure}

\paragraph{Subtype-level trajectories.}
Fig.~\ref{fig:supp-class-recall} resolves development performance by subtype in both task orders.

\section{Evaluation on held-out TCGA patients}
\label{sec:supp-test}
\paragraph{Cohort and separation from development.}
Tab.~\ref{tab:supp-test-cohort} gives the subtype composition of the held-out cohort introduced in main-text Sec.~\ref{sec:datasets}. Patient identity uses the first three TCGA barcode components. We verified the four original split manifests against their archived hashes for both model families. Test membership was identical across all five fold columns and had no patient overlap with the full 1,523-patient development pool. Labels were consistent within patients, slide identifiers were unique across stages, and every included slide had its original cache. All 394 feature files passed dimension, coordinate, finite-value and SHA-256 checks. Cache hashes were distinct within Test and from the sealed development inventory. The audit retained the complete cohort.

\begin{table}[t]
\caption{Held-out cohort shared by all methods and conditions.}
\label{tab:supp-test-cohort}\centering\small
\setlength{\tabcolsep}{3pt}
\begin{tabular}{@{}lrr@{}}\toprule
Subtype & Patients & Slides\\\midrule
LUAD & 21 & 22\\ LUSC & 15 & 15\\ IDC & 138 & 145\\ ILC & 48 & 49\\
KIRC & 101 & 103\\ KICH & 22 & 24\\ EAC & 13 & 13\\ ESCC & 21 & 23\\\midrule
Total & 379 & 394\\
\bottomrule\end{tabular}
\end{table}

\paragraph{Frozen models and inference.}
Each method contributes 16 previously trained trajectories from folds 2--5, forward/reverse orders and seeds 60/61, with four terminal checkpoints per trajectory. We verified all 128 checkpoint receipts and model hashes. No training, calibration, threshold fitting or model selection used these test patients. Before test inference, a development check on 64 patients and 103 slides reproduced all 7,680 archived patient probability entries exactly. Inference then used the same scoring code, model-specific input preparation and original GigaPath cache. Bag selection and inference follow the main-text protocol. Each stage included all test patients from the subtypes introduced by that stage.

Tab.~\ref{tab:supp-test-stages} reports macro-recall across stages.

\begin{table}[!htbp]
\caption{Held-out macro-recall (\%), eight conditions per order. E: EviSlide-MIL; A: ABMIL + Replay/KD. Stage-wise evaluation covers the complete set of seen classes.}
\label{tab:supp-test-stages}\centering\small
\setlength{\tabcolsep}{3pt}
\begin{tabular}{@{}lrrrr@{}}\toprule
 & \multicolumn{2}{c}{Forward} & \multicolumn{2}{c}{Reverse}\\
\cmidrule(lr){2-3}\cmidrule(lr){4-5}
Stage & E & A & E & A\\\midrule
1 & 94.05 & 91.61 & 99.70 & 98.56\\
2 & 88.85 & 76.34 & 95.78 & 94.54\\
3 & 89.65 & 83.04 & 91.24 & 87.30\\
4 & 79.16 & 71.77 & 83.19 & 72.87\\
\bottomrule\end{tabular}
\end{table}

\paragraph{Bootstrap implementation.}
The estimands and resampling count follow main-text Sec.~\ref{sec:datasets}. We used NumPy PCG64 with seed 20261115. Each subtype-stratified patient draw is shared across both methods, all conditions and applicable stages. Each draw recomputes macro-F1, recalls and the peak-recall term in forgetting before equally averaging conditions and forming paired differences. Percentile intervals condition on the fitted models and fixed class composition. The same draws yield the order-specific estimates in Tab.~\ref{tab:supp-test-orders}.

\begin{table}[H]
\caption{Held-out paired differences (EviSlide-MIL minus ABMIL + Replay/KD), with patient-bootstrap 95\% CIs. Means average eight conditions per order. F1 is a percentage; forgetting and paired differences are measured in percentage points.}
\label{tab:supp-test-orders}\centering\small
\setlength{\tabcolsep}{3pt}
\begin{tabular}{@{}llrr@{}}\toprule
Order & Metric & EviSlide & ABMIL\\\midrule
Forward & F1 & 72.93 & 61.40\\
 & \multicolumn{3}{l}{$\Delta=+11.53\ [8.05,15.09]$}\\
 & Forgetting & 19.44 & 30.52\\
 & \multicolumn{3}{l}{$\Delta=-11.08\ [-15.21,-7.41]$}\\\midrule
Reverse & F1 & 78.75 & 65.14\\
 & \multicolumn{3}{l}{$\Delta=+13.61\ [10.47,16.96]$}\\
 & Forgetting & 13.73 & 27.49\\
 & \multicolumn{3}{l}{$\Delta=-13.76\ [-18.02,-9.91]$}\\
\bottomrule\end{tabular}
\end{table}

\paragraph{Additional uncertainty and subtype results.}
The marginal 95\% intervals for final macro-F1 were 72.25--79.22\% for \method{} and 60.43--65.85\% for ABMIL with replay/KD; main-text Sec.~\ref{sec:heldout-results} reports their pooled means and paired effects. Tab.~\ref{tab:supp-test-subtypes} provides the forward-order subtype recalls.

\begin{table}[H]
\caption{Forward held-out subtype recall (\%), averaged over eight conditions. E: EviSlide-MIL; A: ABMIL + Replay/KD. Dashes indicate subtypes not yet introduced at that stage.}
\label{tab:supp-test-subtypes}\centering\footnotesize
\setlength{\tabcolsep}{1.5pt}
\begin{tabular}{@{}llrrrrrrrr@{}}\toprule
Stage & & LUAD & LUSC & IDC & ILC & KIRC & KICH & EAC & ESCC\\\midrule
 1 & E & 96.43 & 91.67 & -- & -- & -- & -- & -- & --\\
  & A & 98.21 & 85.00 & -- & -- & -- & -- & -- & --\\
 2 & E & 84.52 & 90.83 & 87.59 & 92.45 & -- & -- & -- & --\\
  & A & 60.71 & 63.33 & 95.11 & 86.20 & -- & -- & -- & --\\
 3 & E & 76.79 & 91.67 & 86.68 & 91.67 & 92.82 & 98.30 & -- & --\\
  & A & 51.19 & 74.17 & 88.50 & 92.19 & 93.32 & 98.86 & -- & --\\
 4 & E & 52.38 & 30.00 & 78.44 & 91.67 & 92.57 & 98.86 & 97.12 & 92.26\\
  & A & 33.93 & 6.67 & 57.61 & 91.67 & 92.33 & 98.86 & 99.04 & 94.05\\
\bottomrule\end{tabular}
\end{table}

Tab.~\ref{tab:supp-test-subtypes-reverse} gives the corresponding reverse-order outcomes.

\begin{table}[H]
\caption{Reverse held-out subtype recall (\%), averaged over eight conditions. E: EviSlide-MIL; A: ABMIL + Replay/KD. Dashes indicate subtypes not yet introduced at that stage.}
\label{tab:supp-test-subtypes-reverse}\centering\footnotesize
\setlength{\tabcolsep}{1.5pt}
\begin{tabular}{@{}llrrrrrrrr@{}}\toprule
Stage & & LUAD & LUSC & IDC & ILC & KIRC & KICH & EAC & ESCC\\\midrule
 1 & E & -- & -- & -- & -- & -- & -- & 100.00 & 99.40\\
  & A & -- & -- & -- & -- & -- & -- & 97.12 & 100.00\\
 2 & E & -- & -- & -- & -- & 93.07 & 97.73 & 92.31 & 100.00\\
  & A & -- & -- & -- & -- & 94.55 & 97.16 & 89.42 & 97.02\\
 3 & E & -- & -- & 92.03 & 89.32 & 92.45 & 97.73 & 83.65 & 92.26\\
  & A & -- & -- & 95.65 & 85.42 & 90.10 & 99.43 & 74.04 & 79.17\\
 4 & E & 92.26 & 81.67 & 81.70 & 90.36 & 88.99 & 98.30 & 76.92 & 55.36\\
  & A & 89.29 & 82.50 & 62.32 & 86.72 & 84.41 & 100.00 & 63.46 & 14.29\\
\bottomrule\end{tabular}
\end{table}

\paragraph{Source data.}
The archived \emph{p0\_test\_analysis.json} records all 32 method-specific trajectories, 128 stage summaries, subtype recalls, per-class forgetting, confusion matrices and paired intervals. Provenance fields bind the checkpoint lock, audited test manifest, prediction artifact and analysis code by SHA-256. An independent calculation from the saved patient predictions reproduced every condition's metrics and the pooled and order-specific bootstrap intervals. The archived aggregate contains no patient identifiers. The project CSVs cover development experiments; the held-out JSON and patient-level predictions are not included in this source package.

\balance
\end{document}